\documentclass[11pt,a4paper]{article}
\usepackage{times,latexsym}
\usepackage{url}
\usepackage[T1]{fontenc}
\usepackage{graphicx}
\usepackage{multirow}
\usepackage[most]{tcolorbox}
\usepackage{tabularx}
\usepackage{booktabs}

\newcolumntype{R}{>{\raggedleft\arraybackslash}X}
\newcolumntype{L}{>{\raggedright\arraybackslash}X}

\usepackage[acceptedWithA]{tacl2018v2}
\usepackage{xspace,mfirstuc,tabulary}

\newif\iftaclinstructions
\taclinstructionsfalse % AUTHORS: do NOT set this to true
\iftaclinstructions
\renewcommand{\confidential}{}
\renewcommand{\anonsubtext}{(No author info supplied here, for consistency with
TACL-submission anonymization requirements)}
\newcommand{\instr}
\fi

\newtcolorbox{mybox}[2][]{%
  attach boxed title to top center
               = {yshift=-8pt},
  colback      = black!5!white,
  colframe     = black!75!black,
  fonttitle    = \bfseries,
  colbacktitle = black!85!black,
  title        = #2,#1,
  enhanced,
}

\definecolor{ao}{rgb}{0.0, 0.5, 0.0}
\definecolor{carnelian}{rgb}{0.7, 0.11, 0.11}

\definecolor{CMpurple}{rgb}{0.6,0.18,0.64}

\title{Towards Ecologically Valid Research on Language User Interfaces} 

\author{
 Harm de Vries$^1$ \quad Dzmitry Bahdanau$^1$ \quad Christopher Manning$^{2,3}$\\
 \\
 $^1$Element AI \quad $^2$Stanford University \quad $^3$CIFAR Fellow\\
 \\
  {\sf \{harm.de-vries,dzmitry.bahdanau\}@elementai.com} \\
  {\sf manning@stanford.edu}
}

\date{}

\begin{document}
\maketitle
\begin{abstract}
% \cmm{Is NLI the best acronym? I agree things aren't very standardized, but even beyond the ``Natural Language Inference'' confusion, I think LUI is more common; cf. https://en.wikipedia.org/wiki/Natural-language\_user\_interface or https://explosion.ai/blog/natural-user-interface}
Language User Interfaces (LUIs) could improve human-machine interaction for a wide variety of tasks, such as playing music, getting insights from databases, or instructing domestic robots. In contrast to traditional hand-crafted approaches, recent work attempts to build LUIs in a data-driven way using modern deep learning methods. To satisfy the data needs of such learning algorithms, researchers have constructed benchmarks that emphasize the quantity of collected data at the cost of its naturalness and relevance to real-world LUI use cases. As a consequence, research findings on such benchmarks might not be relevant for developing practical LUIs. The goal of this paper is to bootstrap the discussion around this issue, which we refer to as the benchmarks' low \emph{ecological validity}. To this end, we describe what we deem an ideal methodology for machine learning research on LUIs and categorize five common ways in which recent benchmarks deviate from it. We give concrete examples of the five kinds of deviations and their consequences. Lastly, we offer a number of recommendations as to how to increase the ecological validity of machine learning research on LUIs. 
\end{abstract}

\section{Introduction}
In 1991, cognitive scientist Susan E. Brennan wrote the following introduction for one of her papers~\cite{Brennan1991ConversationComputers}:
\begin{quote}
\textit{Why is it that natural language has yet to become a widely used modality of
human/computer interaction? Visionaries seem to have no difficulty imagining a future where we'll be able to talk to software applications -- or even computer agents -- in plain English. And yet the only exposure large numbers of users have had to such interfaces has been through limited question answering systems and keyword interfaces to adventure
games.}
\end{quote}
Nearly three decades later, her observation still holds: Language User Interfaces (LUIs) only play a limited role in our daily interaction with machines.
The recent technological advances in Natural Language Processing (NLP)~\cite{Sutskever2014SequenceNetworks,Bahdanau2015NeuralTranslate,Vaswani2017AttentionNeed,Devlin2019BERT:Understanding}, somewhat surprisingly, have not yet moved the needle in terms of LUI adoption. This motivates us to discuss how academic research on LUIs can be made more aligned with the goal of developing practical LUIs.

%We adopt a human-centric perspective
We are interested in language user interfaces that enhance human capabilities. Specifically, we focus on interfaces that 
%have a clear user need and
support performing a useful and concrete task, such as searching for information in large collections of documents, booking flights, getting insights from statistical data or instructing domestic robots. In the dialogue literature, these systems are referred to as goal-oriented~\citep{Serban2018AVersion} because they facilitate the completion of an unambiguous task, often within a small number of interactions. We distinguish this from the line of work on social chatbots (also known as chit-chat systems)~\cite{Ram2018ConversationalPrize,Zhou2020TheChatbot,Adiwardana2020TowardsChatbot} whose purpose is to engage and entertain users. 

\begin{figure*}[t]
    \centering
    \includegraphics[width=\linewidth]{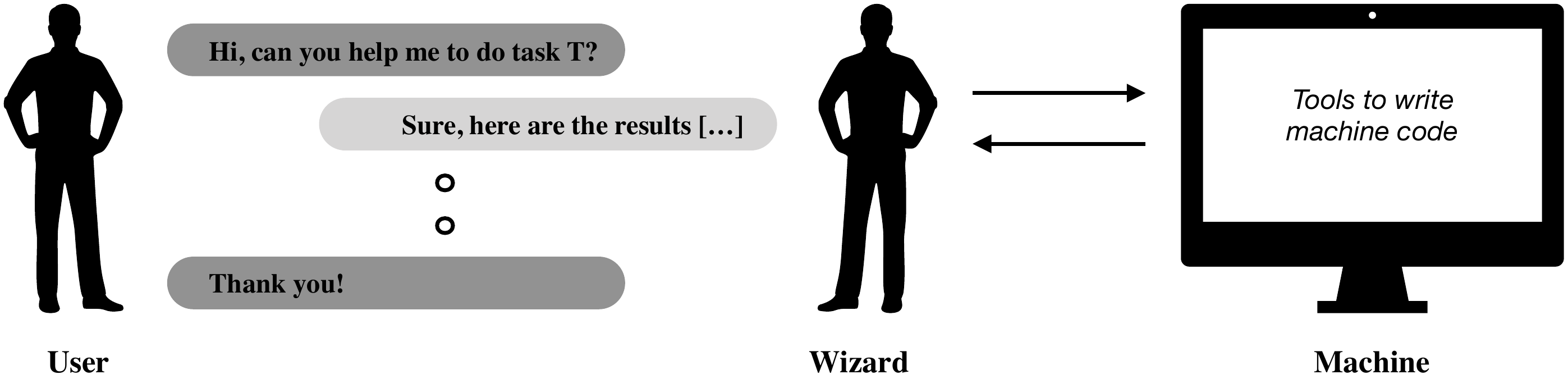}
    \caption{An overview of a Wizard-of-Oz (WoZ) data collection method where a user (on the left) converses with a wizard so as to accomplish their goal. The wizard interprets the user's natural language instructions and translates them to code that the machine can understand (e.g., SQL queries). }
    \label{fig:woz_overview}
\end{figure*}

At the time of Brennan's writing, developing language user interfaces was done by symbolic AI engineers, who analyzed the problem domain and designed linguistic rules for mapping utterances onto formal meaning representations (see, e.g., a survey of early language user interfaces to databases by \citet{Androutsopoulos1995NaturalIntroduction}). While the rule-based paradigm is still widespread (see, e.g., a more recent review by \citet{affolter_comparative_2019}), its scalability is limited by the large amounts of expert labor needed to develop, maintain and adapt such systems. We therefore focus our analysis on the Machine Learning (ML) approach, in which the bulk of knowledge about natural language is entered as example utterances or dialogues. The use of data instead of expert labor promises better scalability and flexibility, but a key assumption underlying these hopes is that the data is available and appropriate.

%\cm{I think we want to contrast 3 situations: (i) Applied scientists working on, say, Siri, Alexa, or Google web question answering who \emph{do} have tons of ecologically valid data (but generally don't publish or provide resources to the research community; (ii) HCI work which is generally ecologically valide but to achieve this has focused on small user studies of commercial LUIs, e.g., \citep{luger2016like,cowan2017what}, and (iii) research work (both academic and in industry research labs such as FAIR). It is the tasks and datasets of this third group that are the focus of this paper.} 

%Ideally,  the  data  used  for  training  and  evalu-ating  the  learned  LUIs  should  reflect  the  intentsand linguistic phenomena found in real-world applications and be of reasonable size to accommo-date modern data-intensive methods. In a few spe-cific industrial settings such data may be readily available as logs of users interacting with an ex-isting  service,  as  e.g.   Siri  or  Google  Assistant.Such datasets are rarely publicly available due totheir commercial value.   Moreover,  for LUI use-cases  without  a  widely  used  working  prototypesuch  data  most  likely  does  not  exist  at  all.   Forthese  diverse  practical  and  research  settings  thetwo requirements of data quality (and in particular its representativeness) and quantity are hard toreconcile. 

Ideally, the data used for training and evaluating the learned LUIs should reflect the intents and linguistic phenomena found in real-world applications and be of reasonable size to accommodate modern data-intensive methods. In certain industrial  settings such data might be readily available as logs of users interacting with an existing interface, e.g. Siri or Google Assistant. Such datasets are rarely publicly available both due to customer data privacy needs and their commercial value. % Moreover, for LUI use cases without a widely used working prototype such data may not exist at all. 
For the broader research community, on the other hand, the two requirements of data quality (and in particular its representativeness) and quantity are hard to reconcile.

Earlier literature features data collection efforts of exceptional execution quality, in which researchers attempted to closely simulate the LUI's anticipated use-case.
Some of them were run as user studies~\citep{Grosz1974TheDialogs,kelley_iterative_1984,Dahlback1993WizardStudies,Carbonell1983DiscourseInterfaces}, whereas others aimed to collect data for automatic evaluation purposes~\citep{Hemphill1990TheCorpus,dahl_expanding_1994}. In both cases, the methods employed to achieve this quality were expensive and hard to scale up. The vast majority of the collected corpora contains anywhere from tens to hundreds of utterances, which is hardly enough for deep-learning-based approaches. 
% - a simulation must be built, quoting NLT' 90, a "construction project"
% - representative users must be found and paid
% - qualified wizard must be found and paid

In pursuit of more data, many recent benchmarks opted for cheaper and more scalable methods. For example, it is common these days to use artificial tasks with no naturalistic counterparts or to work with crowd workers that are not representative of the target user population. 
% Here, let's do not put guilt on a few chosen papers. Praising something can be done almost without context. Cricizing must be done carefully :) 
It is unclear what the consequences of these compromises are for the transferability of research findings. In particular, one can wonder to which extent improvements on these benchmarks translate into more useful language user interfaces.

The questions that we pose above correspond to the notion of external, and more specifically, \emph{ecological validity} from the psychology literature. The conclusions of an externally valid experiment should hold outside the context of that study~\cite{bronfenbrenner1977toward,Brewer2014ResearchValidity}. For psychological studies it often indicates whether a causal effect holds up across different populations, environments, or stimuli. Ecological validity is a special case of external validity, specifying the degree to which findings generalize to naturally occurring scenarios. % Thus, research conducted in situations that are likely to happen in real-life is said to have high ecological validity. The key strength of such studies is that it ensures that generated insights are practically relevant and useful.
The key strength of studies with high ecological validity is that they generate insights that are practically relevant and useful. 
% Dima: here is your HCI mention
Such studies on LUIs are commonly found in the Human Computer Interaction (HCI) community, e.g. by conducting interviews with real-world users of commercial personal assistants~\cite{luger2016like, cowan2017what}. 
%Achieving this level of ecological validity in mainstream 

% By contrast, internal validity concerns the validity of the conclusion within the context of the study.

% In this paper, we wish to facilitate discussions on the ecological validity of LUI research benchmarks. To this end, we sketch what we think is an ecologically valid research methodology and review how recently proposed benchmarks deviate from it. We find five common issues---synthetic language, artificial tasks, not working with prospective users, the use of scripts and/or priming, and single-turn interfaces---and show through concrete examples how these issues limit the benchmarks' ecological validity. The rest of this paper is organized as follows. To make the paper's scope concrete, we discuss several important LUI usecases in Section 2. In Section 3, we describe in broad strokes our position as to how ecologically valid research on LUI should be conducted. In Section 4, we categorize five common deviations from the proposed ideal methodology and provide examples of their consequences. We discuss other ecological validity concerns in Section 5 and offer recommendations as to how to increase the ecological validity of machine learning research on LUIs in Section 6. 

With this paper, we wish to encourage discussions on the ecological validity of LUI research benchmarks. We first discuss several important LUI usecases in Section 2 to make the paper's scope concrete. In Section 3, we sketch what we think is an ecologically valid research methodology for how valid research on LUI should be conducted. We then review in Section 4 how recently proposed benchmarks deviate from it and find five common issues---synthetic language, artificial tasks, not working with prospective users, the use of scripts and/or priming, and single-turn interfaces---and show through concrete examples how these issues limit the benchmarks' ecological validity. We discuss other ecological validity concerns in Section 5 and conclude with recommendations as to how to increase the ecological validity of machine learning research on LUIs in Section 6. 

\section{Examples of Language User Interfaces}

% \cm{I think some of the taxonomy here could be better, and a few more examples mentioned. It seems like a big division is whether there is conversion to a formal language or things are found in text. The former groups database queries with booking services or buying goods from web APIs, while the latter groups web QA with things like customer support QA. I think you should more generally mention getting information from Web APIs (e.g., weather, Yelp restaurant information, movie times), tasks via web APIs (purchasing products, tickets, or services), and calendar assistance -- that's a key NL use case. Also, as well as the somewhat futuristic domestic robots, mentioning domestic appliances, using NL to control TVs, smart home devices, and of course Siri or Google Assistant on your phone.}

Before we discuss the notion of ecological validity, we will make the concept of language user interfaces more tangible by listing a number of prominent use cases. Note that we intentionally focus on the end user applications and not on the underlying technologies or the corresponding academic ``tasks'', such as question answering and semantic parsing, which are more commonly referred to in academic papers. It is the close connection to one of such real-world use cases that makes a task or benchmark ecologically valid. 

% % - Personal Assistants
%    - Siri, google assistants
% - Customer support assistants 
% - Business Intelligence Assistants
% - Assistants for the visually impaired
% - Domestic appliances and domestics robots

\paragraph{Personal assistants} 
LUIs could aid people in the organisation of their daily lives. For example, such personal assistants can help with obtaining weather forecasts, managing calendar events, and reserving restaurant tables. Google Assistant and Siri are two well-known examples of LUIs that are aiming to provide these services.

\paragraph{Assistants for the visually impaired} LUIs could aid blind people in overcoming many of their daily challenges. One can, for example, think of an application where visually impaired people could take pictures of their immediate surroundings and ask targeted questions about its content~\cite{Gurari2018VizWizPeople}. Such assistants could enable them to identify objects, read text labels, or obtain other information that is usually only available to people with good eyesight.

% Should we talk about the difference between relational databases and knowledge graphs? 

\paragraph{Customer support assistants} LUIs could provide customer support for purchasing goods and services. Such assistants could, for example, guide the customer through the buying process through a chat-based interface. They can also answer detailed questions about the service provider's policies, going beyond the lengthy list of Frequently Asked Questions (FAQ).

% \cite{Hemphill1990TheCorpus,Wen2016ASystem}. 

\paragraph{Assistants for business analysts} LUIs could help analysts obtaining insights into business processes. While this task usually requires writing SQL queries for relational databases or navigating graphical dashboards, LUIs can improve the user experience by enabling natural language requests. 

\paragraph{Domestic appliances and robots} LUIs could aid our interaction with domestic appliances and robots. For example, they would enable controlling TVs or other appliances through natural language instructions. In the longer term, LUIs could also be helpful for interacting with physical robots, e.g. to instruct them to iron shirts or clean the floor. 

% This line of work on instruction-following tasks has recently gained traction through the development of virtual environments for robotics research~\cite{Das2018EmbodiedAnswering,Anderson2018Vision-and-LanguageEnvironments,Chevalier-Boisvert2018BabyAI:Learning}. The input-output examples for these tasks consists of natural language utterances and corresponding \emph{action sequences} to be executed by the robot.

% \paragraph{Other natural language interfaces} 

\section{Ecologically Valid Research}\label{sec:goldstandard}
By the very definition of the concept, ecologically valid research on LUIs should strive to build LUIs that people would enjoy or benefit from using in their personal or professional life. It should thus start with identifying a population of people $P^{\mathcal{T}}$ in need of assistance with task $\mathcal{T}$.
Moreover, the developed LUI for this task needs to be more valuable or usable than available alternatives. For example, because users are unable to complete the task with the current interface or would be more satisfied with a language user interface. 
\begin{table}[t]
\begin{mybox}[colback=white]{An Ecologically Valid Research Procedure}
\begin{enumerate}
\item Identify a population of users $P^{\mathcal{T}}$ who would benefit from a language user interface to perform a task $\mathcal{T}$. The constructed LUI should increase the user's productivity in task $\mathcal{T}$ compared to alternative interfaces. 
%\item Identify a population of wizards $P^{\mathcal{T}}_{wizard}$ who are able to perform the task $\mathcal{T}$ (with help of another graphical interface).
\item Collect conversations and corresponding programs/actions through a Wizard-of-Oz simulation of performing task $\mathcal{T}$.
\item Train a model
\item Assess how satisfied the user is with the trained model through a $P^{\mathcal{T}}$-in-the-loop evaluation. 
\end{enumerate}
\end{mybox}
\caption{Summary of the proposed ecologically valid research procedure on LUIs.}
\label{table:goldstandard}
\end{table}

Identifying the user and the task is step 1 out of 4 in the ideal methodology that we propose here (see Table \ref{table:goldstandard}). Step 2 is to collect data to train the system. For machine learning it is of utmost importance to gather data under conditions that are similar to the deployment setting. Yet the exact deployment setting cannot be simulated until the system is trained and deployed. 

To bootstrap out of this chicken and egg problem, the yet-to-be-built LUI can be simulated by a ``wizard''. The wizard translates the user's instructions to programs (e.g., SQL queries) or actions that the machine can execute, often with the help of specifically designed tools for the task (see Figure \ref{fig:woz_overview}). The described approach is often referred to as Wizard-of-Oz (WoZ)~\cite{kelley_iterative_1984,Fraser1991SimulatingSystems,Maulsby1993PrototypingOz}. Ideally, in a WoZ simulation the user should think that they interact with a machine and not know there is a human ``behind the curtain''. The argument for maintaining the illusion is that people adjust their language to the characteristics of the listener~\cite{Shatz1973TheListener}, implying that users interact differently with machines than with human interlocutors. Work from the 80s and 90s puts a lot of emphasis on this aspect of the simulation~\cite{kelley_iterative_1984,Dahlback1993WizardStudies}. However, the use of a human wizard is not concealed in \citet{sinha2002embarking} and it is unclear if, in 2020, people could be easily convinced that they are interacting with a machine despite the unavoidable long response times (for it takes time for the wizard to execute what the user wants and possibly also respond). For this reason, we will not view the user's awareness of simulation as a deviation, even though strictly speaking it is one. 

Once the WoZ setup is constructed, one collects a large number of WoZ interactions and records the conversations as well as the programs or actions created by the wizard. Using this dataset, one trains a model to interpret the natural language conversations and predict the programs or actions of the wizard. Finally, one assesses how satisfied the user is with the resulting model and compares it to a traditional interface or a competing model. Ideally, this process measures several aspects of human satisfaction through a human-in-the-loop evaluation with users coming from $P^{T}$.
% Training and evaluation constitute the third and final step in the proposed methodology.
% We will refer to this guideline as an \textit{ecologically valid} procedure and briefly summarize it in Table \ref{table:goldstandard}.  

\begin{table}[t!]
\small \begin{tabularx}{\linewidth}{lL}
\textit{User:} & \textit{good morning}\\
\textit{Wizard:} & \textit{hello what can i help you with today}\\
\textit{User:} & \textit{can you book a table in a moderate price range with british cuisine for eight people in rome}\\
\textit{Wizard:} & \textit{i'm on it}\\
\textit{User:} & \textit{SILENCE}\\
\textit{Wizard:}  & \textit{ok let me look into some options for you}\\
\textit{User:} & \textit{SILENCE}\\
\textit{Wizard:} & \textit{api\_call british rome eight moderate}\\
\textit{User:} & \textit{instead could it be for six people}\\
\textit{Wizard:} & \textit{sure is there anything else to update}\\
\textit{User:} & \textit{instead could it be in bombay}\\
\textit{Wizard:} & \textit{sure is there anything else to update}\\
\textit{User:} & \textit{instead could it be with italian food}\\
\textit{Wizard:} & \textit{sure is there anything else to update}\\
\textit{User:} & \textit{actually i would prefer in a expensive price range} \\
\textit{Wizard:} & \textit{sure is there anything else to update}\\
\textit{User:}  & \textit{no}\\
\textit{Wizard:} & \textit{ok let me look into some options for you}\\
\textit{User:}  & \textit{SILENCE}\\
\textit{Wizard:} & \textit{api\_call italian bombay six expensive}\\
\textit{User:} & \textit{thanks}\\
\textit{Wizard:} & \textit{you're welcome}\\
\end{tabularx}
\caption{Example of the refine API task from the dialog-bAbI dataset~\cite{Bordes2016LearningDialog}. The repeated update of the restaurant reservation is repetitive and lacks the diversity found in human dialogues.}
\label{tab:bAbI}
\end{table}

\section{Deviations}\label{sec:limitations}

A large majority of recent pure research projects from the ML and NLP communities do not align with the proposed notion of an ecologically valid research procedure. This section describes five common issues---synthetic language, artificial task, not working with prospective user, scripts and priming, single-turn interfaces---and points out their limitations through concrete examples. For many benchmarks, the lack of ecological validity comes from multiple factors which are often hard to disentangle. For that reason, we pick a few example projects that best illustrate the potential impact of this deviation from the ideal data collection methodology. 

\subsection{Synthetic language}\label{sec:syntheticlanguage}
Perhaps the most obvious deviation is to dismiss any form of data collection and instead work with synthetic language. The key difficulty in designing a synthetic language is to obtain broad linguistic coverage while maintaining the natural aspect of language. Some projects intentionally keep the language simple and coverage low. The BabyAI project~\cite{Chevalier-Boisvert2019BabyAI:Loop} defines a context-free grammar to generate simple instructions such as 
\begin{quote}\textit{open the yellow door, then go to the key behind you}.
\end{quote}
While the BabyAI grammar can generate a large number of instructions, its vocabulary is small and features only a few dozen words.  In addition, they need to impose restrictions on the use of ``and'', ``then'', and ``after you'' connectors to maintain the readability of the instructions. In general, it is important for grammar-based approaches to carefully limit the operators that can lead to combinatorial explosion, as these are often the source of unnatural utterances. For example, some questions in the Compositional Freebase Questions (CFQ) dataset~\cite{Keysers2019MeasuringData} are hard to understand because of the conjunction of many noun or verb phrases:
\begin{quote}
\textit{Did Patrick Scully's sibling marry Carolyn Zeifman, influence Tetsuo II: Body Hammer's art director, director, and executive producer, and influence Christophe Gans?}
\end{quote} 
Especially for larger domains it becomes increasingly difficult and tedious to ensure the readability of all questions or instructions (see, e.g., the effort by \citet{hudson_gqa_2019}). 
% or
% \begin{quote}
% \textit{Which screenwriter did Louis Garrel's spouse's sibling marry and Karine Silla and Jacqueline Bissett marry?}
% \end{quote} 

Long natural-looking questions or dialogues often feature anaphoric references, e.g., the pronoun ``them'' in the following instruction: 
\begin{quote}
\textit{pick up my shoes, then bring \textbf{them} to the living room}.
\end{quote}
% \cmm{There \emph{was} a lot of work earlier on referring expression generation, e.g., see https://www.aclweb.org/anthology/J12-1006.pdf -- so we should acknowledge that somehow.} 
Generating synthetic data containing a wide variety of such references has been studied but remains challenging \citep{krahmer_computational_2012}. The existing synthetic datasets  feature only very restricted use of pronouns and are usually template-based. %In contrast to context-free grammars, template-based methods usually include a limited number of pronouns . 
For example in CLEVR~\cite{Johnson2017Clevr:Reasoning}, the authors manually write templates for each high-level intent, which contain a number of slots that are filled during instantiation of the question. %or instruction. 
A drawback of designing templates is that it is labor-intensive and only features relatively few pronouns (namely, the ones that the authors wrote). Producing anaphoric references in a conversational setting is even more challenging as they might refer back to previous dialogue turns (e.g., the pronoun ``them'' can refer to the object ``shoes'' in the previous turn). Most synthetic dialogue datasets write templates for each dialogue act independently, which can lead to conversations in which the dialogue acts are not ``smoothly'' connected. See, for example, the dialog of the bAbI dataset~\cite{Bordes2016LearningDialog} in Table \ref{tab:bAbI}, in which the repeated update of the restaurant reservation is repetitive and unnatural. 
% \cm{Another good example to discuss would be \citet{campagna2019genie}, where a synthetic grammar is used to produce a lot of training data, but it is supplemented with Turker paraphrases because otherwise you just don't see enough linguistic variety to produce a system that works in the world.}

To summarize, developing a synthetic language that is both natural-looking and covers all necessary linguistic phenomena is highly challenging. Findings on synthetic benchmarks might therefore not be representative of progress on practically relevant LUIs. 

% CLEVR example: 
% {Were [It’s Not About the Shawerma], [The Fifth Wall], [Rick’s Canoe], [White Stork Is Coming], and [Blues for the Avatar] executive produced, edited, directed, and written by a screenwriter’s parent?7! Yes}

% Linguistic coverage and . 
% Which screenwriter did Louis Garrel's spouse's sibling marry and Karine Silla and Jacqueline Bissett marry

% Do we have an example of a model that worked well on synthetic data but did not transfer to natural language?

\begin{figure}[t]
\begin{tabular}{lr}
\multicolumn{2}{@{}c@{}}{\includegraphics[width=\linewidth]{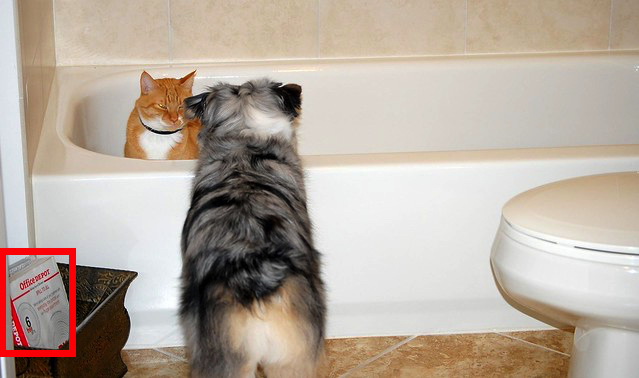}}\\
\textit{Is it an animal?} & \textcolor{carnelian}{No}\\
\textit{Is it white?} & \textcolor{ao}{Yes}\\
\textit{Is it only on the right half of the picture?} & \textcolor{carnelian}{No}\\
\textit{Is the cat sitting in it?} & \textcolor{carnelian}{No}\\
\textit{Are there words on it?} & \textcolor{ao}{Yes}\\
\end{tabular}
\caption{An example of the ``artificial task'' deviation: the GuessWhat game~\cite{DeVries2017GuesswhatDialogue} in which users ask yes-no questions in order to find the hidden object (highlighted by the red bounding box). %\cm{Could there be a red box or similar instead of green mask -- or else the ``Is it white?'' question is quite confusing.}
}
\label{fig:guesswhat}
\end{figure}

\subsection{Artificial task} \label{sec:artificial_task}
\begin{figure}[t]
\begin{tabular}{@{}l@{}}
\includegraphics[width=\linewidth]{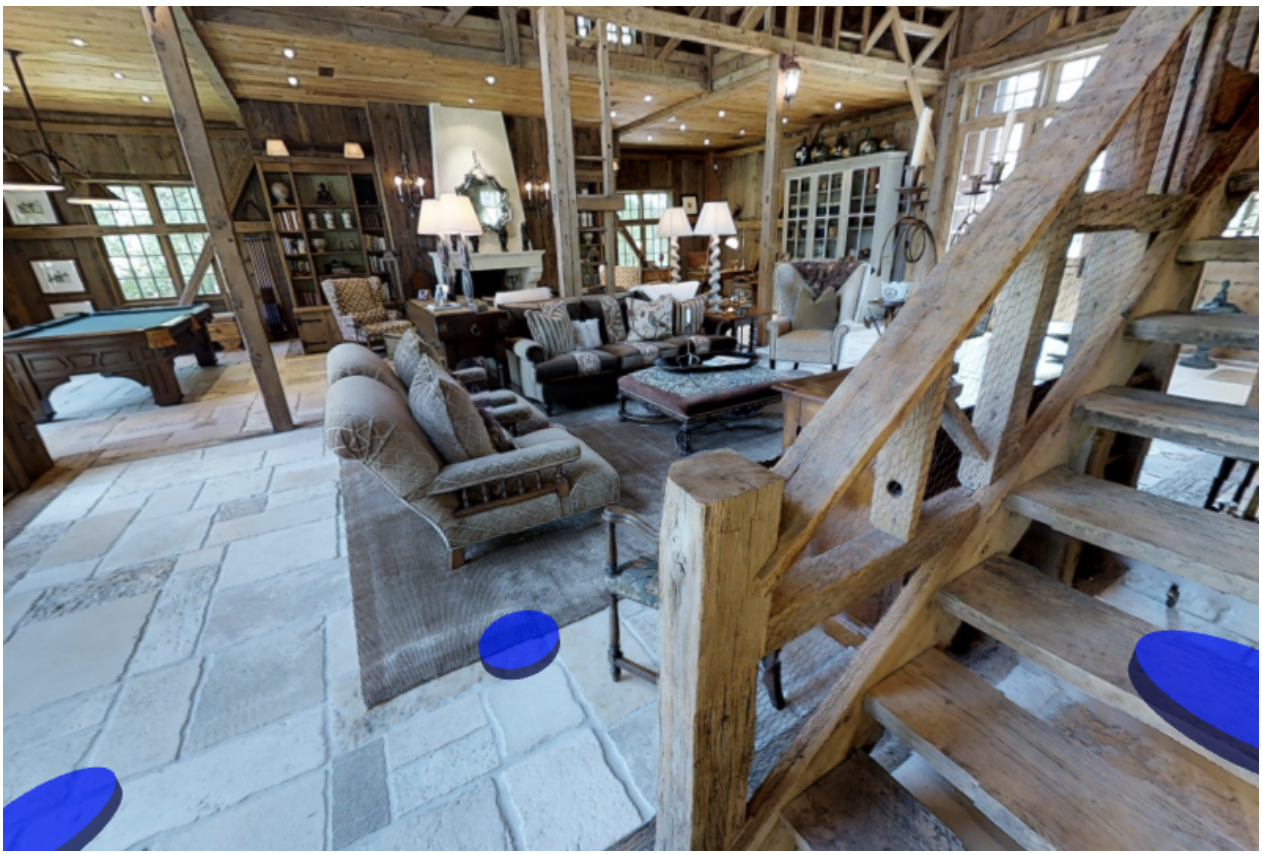}\\
\small\textit{\textbf{Instruction:} Head upstairs and walk past a piano}\\ 
\small\textit{through an archway directly in front. Turn right}\\ 
\small\textit{when the hallway ends at pictures and table. Wait}\\ 
\small\textit{by the moose antlers hanging on the wall.}
\end{tabular}
\caption{Another example of the ``artificial task'' deviation: the Vision-and-Language benchmark~\cite{Anderson2018Vision-and-LanguageEnvironments} proposes a LUI task for robot navigation. }
\label{fig:vnl}
\end{figure}

Crafting custom artificial tasks (games) for research purposes is another common deviation from the ideal procedure. Such tasks may be appealing in that they require advanced linguistic human-computer interaction, and the associated data collection efforts often yield diverse and interesting data. Nevertheless, we deem it problematic that these tasks do not correspond to or even resemble a practically relevant LUI setting. For example, Room2Room~\cite{Anderson2018Vision-and-LanguageEnvironments} proposes a LUI task to let robots navigate to (random) locations in the Matterport 3D environment---see Fig. \ref{fig:vnl} for an example. We expect that robots will be used to e.g., find and pick up objects in a household setting, a task for which navigation is only a subroutine. Human instructions for household tasks are probably more high-level and unlikely to contain detailed information on the navigation task, such as where to turn left\slash right. This mismatch decreases the ecological validity of the Room2Room benchmark.

The vast majority of other artificial tasks are cast as games. One prominent example is the GuessWhat task~\cite{DeVries2017GuesswhatDialogue}, a 20Q inspired game in which the user aims to find a hidden object in an image. The user can ask a series of yes-no questions to the wizard, who can see the hidden object. See Fig. \ref{fig:guesswhat} for an example dialog. Another example is CerealBar~\cite{Suhr2019ExecutingInteractions}, where two agents, a leader and a follower, navigate a toy 3D environment in order to collect a sets of cards. The leader agent has an overview map of the environment but cannot take as many steps as the follower agent. They therefore delegate the collection of some cards to the follower by providing natural language instructions. Similar to GuessWhat, the CerealBar task is an artificially constructed game that is only meaningful within this virtual environment. We categorize such benchmarks as having low ecological validity because we do not think that people would naturally use these LUIs. 

Note that not all game environments are automatically classified as such. Popular game environments, like Minecraft~\cite{Szlam2019WhyMinecraft}, could be an excellent platform for developing LUI tasks with high ecological validity.  It should also be noted that, despite the ecological validity concerns, artificial tasks can still serve as an interesting playground for working on conceptual advances in learning and modelling methods. We believe that they are ill-suited for incremental research, as it is unclear how small improvements will find their way to real applications. 

% \cmm{Would it be better to start with this example and then say that most are games, so one can end the section with the hopeful Minecraft example?} 

\subsection{Not working with prospective users}\label{sec:not_prospective}

One of the most common issues with existing LUI datasets is that the population that would actually benefit from the language user interface rarely participates in the data collection effort. An example of what this can lead to can be found in the context of visual question answering (VQA). This task gained interest from the research community after the release of the VQA dataset~\cite{Antol2015VQA:Answering}, consisting of more than 750K open-ended questions. The contextually rich images in VQA are taken from MS COCO~\cite{Lin2014MicrosoftContext} and natural language questions are gathered on a crowd-sourcing platform via the following set of instructions:
\begin{quote}
\textit{We have built a smart robot. It understands a lot about
images. It can recognize and name all the objects, it
knows where the objects are, it can recognize the scene
(e.g., kitchen, beach), people's expressions and poses, and
properties of objects (e.g., color of objects, their texture).
Your task is to stump this smart robot!
Ask a question about this scene that this smart robot
probably cannot answer, but any human can easily
answer while looking at the scene in the image.}
\end{quote}
Although the VQA task was at least partly inspired by the need to help the visually impaired,\footnote{Taken from the abstract of \citet{Antol2015VQA:Answering}: ``Mirroring real-world scenarios, such as helping the visually impaired, both the questions and answers are open-ended''} questions were not collected from blind people. Instead, human subjects with 20/20 vision were primed to ask questions that would stump a smart robot. As shown by the VizWiz project~\cite{Gurari2018VizWizPeople}, this decision has had a profound impact on the ecological validity of the dataset. Specifically, their case study found that blind people (1) ask questions that are sometimes incomplete and often conversational in nature, (2) start their questions almost always with ``what'' (as opposed to words that narrow the answer space, such as ``how many'' or ``is it''), and (3) frequently formulate questions that require text-reading capabilities (in 21\% of the cases). In addition, blind photographers captured the images using their mobile phone, resulting in many unanswerable questions because of the poor image quality or irrelevant content. Perhaps as a consequence of the differences in the datasets, the authors reported that modern VQA models struggle on the VizWiz dataset, especially when it comes to answering questions that require text-reading capabilities. 

In the context of database QA, Spider~\cite{Yu2018Spider:Task} collected questions from 11 computer science students with proficiency in SQL. For each of the 200 databases, they were instructed to write 20--50 questions so as to cover a number of SQL patterns. The students did not have the intention to find information in the database, resulting in questions that might not align with the user population. Looking at the data, we observed that some questions are quite literal translations of the SQL query, sometimes explicitly referring to column names\footnote{As noted by \citet{Suhr2020ExploringParsing},  questions in other database QA datasets rarely mention the column name.}:
\begin{quote}
\textit{What are the names of the customers who bought product "food" at least once?}
\end{quote}
The SQUAD dataset~\cite{Rajpurkar2016SQuAD:Text} was collected by having human annotators generate questions about Wikipedia articles. Like the Spider project, these crowdworkers had no information need, which makes it unclear if the resulting questions match the ones from users looking for this information. In all three project examples above, the discrepancy between questions collected from wrong or poorly incentivized users versus target users can make trained models much less useful for the target users than automatic evaluation on the ecologically invalid questions would suggest.

% 28 percent unanswerable, 28 text-based reasoning
% Really have the same information-need. VQA 
% Spider, question sampled from different subpopulation. 

\begin{table}[t]
\small
\textbf{Prompt}
\begin{itemize}
    \item \textit{You are looking for a place to dine. The restaurant should serve chinese food and should be in the south}
\end{itemize}
\begin{tabularx}{\linewidth}{lL}
    \hline
    \textit{User:} & \textit{I need a place to dine at in the south that serves chinese. [...] }
\end{tabularx}
\caption{Example of the MultiWOZ dataset \cite{Budzianowski2018MultiWozModelling}, highlighting how users might copy parts of the textual prompt.}
\label{table:copyprompt_multiwoz}
\end{table}

\begin{table}[t]
\small
\textbf{Prompt}
\begin{itemize}\item \textit{You are looking for a place to stay. The hotel should be in the cheap price range and should be in the type of hotel}
    \item \textit{The hotel should include free parking and should include free wifi}
    \item \textit{Once you find the hotel you want to book it for 6 people and 3 nights starting from tuesday}
    \item \textit{If the booking fails how about 2 nights}
    \item \textit{Make sure you get the reference number}
    \end{itemize}
\begin{tabularx}{\linewidth}{lL}
     \hline
    \textit{User:} & \textit{[sic] am looking for a place [sic] to to stay that has cheap price range it should be in a type of hotel }\\
    \textit{Wizard:} & \textit{Okay, do you have a specific area you want to stay in?} \\
    \textit{User:} & \textit{no, i just need to make sure it's cheap. oh, and i need parking} \\
    \textit{Wizard:} & \textit{I found 1 cheap hotel for you that includes parking. Do you like me to book it?}\\
    \textit{User:} & \textit{Yes, please. 6 people 3 nights starting on tuesday.} \\
    \textit{Wizard:} & \textit{I am sorry but I wasn't able to book that for you for Tuesday. Is there another day you would like to stay or perhaps a shorter stay?} \\
    \textit{User:} & \textit{how about only 2 nights.} \\
    \textit{Wizard:} & \textit{Booking was successful. [...] }
\end{tabularx}
\caption{An example from the MultiWOZ dataset \cite{Budzianowski2018MultiWozModelling}, illustrating how the scripts limits the search for a hotel to at most 2 attempts. }
\label{table:multiwoz_unforeseen}
\end{table}

\begin{table}[t]
\small
\textbf{Prompt}
\begin{itemize}
\item \textit{Of all the claims, what was the earliest date when any claim was made?}
\end{itemize}

\begin{tabularx}{\linewidth}{lL}
    \hline
    \textit{User:} & \textit{What days were the claims made?} \\
    \textit{Wizard (SQL):} & \textit{SELECT Date\_Claim\_Made FROM Claims} \\
    \textit{Wizard (Response):} & \textit{There are the dates when all the Claims were made.} \\
    \textit{User:} & \textit{When was the earliest claim made?} \\
    \textit{Wizard (SQL):} & \textit{SELECT Date\_Claim\_Made FROM Claims ORDER BY Date\_Claim\_Made ASC LIMIT 1}\\
    \textit{Wizard (Response):} & \textit{The date when the earliest claim was made is 2016-08-02.} \\
    \textit{User:} & \textit{What day of the week was that claim made?} \\
    \textit{Wizard (Response):} & \textit{Sorry, I cannot find info about this.}\\
    \textit{User:} & \textit{What was the claim ID for the earliest claim?} \\
    \textit{Wizard (SQL):} & \textit{SELECT Claim\_ID FROM Claims ORDER BY Date\_Claim\_Made ASC LIMIT 1} \\
    \textit{Wizard (Response):} & \textit{The claim id of the earliest claim is 563.}\\
\end{tabularx}
\caption{While the CoSQL task~\cite{Yu2019CoSQL:Databases} instructs users to explore the database, we observe that questions often stay close to (parts of) the prompt. }
\label{table:cosql}
\end{table}

\subsection{Scripts and priming} \label{sec:scripts}
% scripting (MultiWOZ, ALFRED, airdialogue??)
% dialogs stay close to textual prompt (CoSQL, SpaRC)

To compensate for the lack of access to potential users and/or to capable wizards, many recent data collection efforts relied on scripts that constrained the flow of human-computer interaction.  For example, \citet{Budzianowski2018MultiWozModelling} collect the MultiWOZ dataset of dialogues for making reservations in hotels, restaurants, etc. For each dialogue, the user is given a script that they are supposed to follow (see Table \ref{table:multiwoz_unforeseen} for an example). The script defines their preferences, such as the type of food and price range of the restaurant, as well as alternatives if their first choice is unavailable. 

The use of scripts can cause ecological validity issues, two of which we discuss below. For one, the diversity of user-wizard interactions is limited by the complexity of the script. In the case of MultiWOZ, for example, the search for the right hotel or restaurant cannot take more than two turns. If it is impossible to realize the first set of preferences (e.g., no hotel is available for 3 nights), the prompt suggests an alternative that is feasible (e.g., to book for 2 nights instead, see Table \ref{table:multiwoz_unforeseen} for the complete prompt). In dialogues where a user must inquire about several reservations, each reservation is always completed before the next one is started. For example, the user cannot reconsider their choice of restaurant based on the train schedule. It is possible (and perhaps even likely) that models trained on MultiWOZ data will have trouble generalizing to interaction scenarios that the scripts did not cover. 

The second, more direct effect that scripts can have on the collected data is that subjects are primed by the specific wording of the script. In the worst case, users directly copy an automatically generated prompt without rephrasing (e.g., in the first utterance of Table \ref{table:multiwoz_unforeseen}: ``[sic] am looking for a place to to stay that has cheap price range it should be in a type of hotel''). In a less severe example, the user rephrases the prompt to be more plausible, but the resulting request is still unnatural and heavily influenced by the automatically generated prompt. For example, instead of saying ``I need a place to dine at in the south that serves chinese'', most people would probably say ``Chinese restaurant'' or ``Chinese food'' (see Table \ref{table:copyprompt_multiwoz} for the complete example).

There are many other examples of LUI datasets whose diversity and ecologically validity may be negatively affected by scripting or textual priming. The ALFRED task~ \cite{Shridhar2020Alfred:Tasks} collects instructions by having the AMT workers annotate videos of a robot acting in simulated environment. The videos are generated by attaining goals with a planner. There are only 7 kinds of goals (such as ``pick \& place'', ``stack \& place'', etc.) and it is unclear if these are representative of the sort of requests that an actual user would want to accomplish. In CoSQL~\citep{Yu2019CoSQL:Databases}, users engage in a dialogue with a conversational database interface that is enacted by a SQL-competent wizard. The users are instructed to explore the DB and also primed by SQL queries coming from the Spider dataset that we discussed earlier. Looking at CoSQL dialogues, we observed that users often asked questions that were close to (or reformulations of) the prompt or its parts (see an example dialogue in Table \ref{table:cosql}), as opposed to performing curiosity-driven data exploration. The proximity of the dialogue to prompts that originate from SPIDER means that ecological validity concerns regarding Spider queries (see Section \ref{sec:not_prospective}) transfer to CoSQL data.
% DIMA: we could also try argue that the conversation coherence may be affected by the fact that user is not by an inner intent (curiosity). But they, in fact, do encourage (monetarily) people to ask inter-related questions 

\subsection{Single-turn interfaces}\label{sec:singleturn} 
Some recent benchmarks are free of the deviations that we have covered so far as they consider real and useful tasks and involve target users in the data collection effort. For example, the Advising dataset~\cite{Finegan-Dollak2018ImprovingMethodology} collects questions about the course information of the University of Michigan from a department's Facebook page.\footnote{Other ones were collected by instructing CS students with knowledge of the database to write questions they might ask in an academic advising appointment.} Other examples are recent open domain QA benchmarks that extract questions from anonymized logs of a search engine, such as the MS MARCO \cite{Bajaj2016MSDataset}, Google Natural Questions \cite{Kwiatkowski2019NaturalResearch} and DuReader \cite{He2018DuReader:Applications} datasets. These benchmarks much more ecologically valid than the ones we discussed earlier in this paper, yet we note that the user is only allowed to ask a single question, i.e., the interaction is single-turn. These projects thus lack the conversational aspect of the proposed notion for an ecologically valid research procedure, despite many web search sessions spanning multiple queries \citep{raman2013toward}.

The importance of multi-turn interactions has been established through several Wizard-of-Oz studies~\cite{Carbonell1983DiscourseInterfaces,Bertomeu2006ContextualExperiment,Dahlback1989EmpiricalInterfaces}, suggesting that there are qualitative differences with single-turn interfaces. In a case study simulating a sales assistant, \citet{Carbonell1983DiscourseInterfaces} reports that users rely on a rich number of dialog phenomena, such as anaphora, ellipses (see Table \ref{table:ellipses} for an example), and meta-linguistic utterances (``I should have said \ldots''). Interestingly, even when users are explicitly instructed to formulate standalone expressions they tend to produce fragmentary utterances. In a database QA setting, \citet{Bertomeu2006ContextualExperiment} argue that users naturally ask a series of thematically related questions when performing information-seeking tasks. By analyzing a small corpus of QA conversations, they confirm that a large number of questions (~36\%) are indeed context-dependent. These empirical studies suggest that dialog is the preferred mode of interaction for most LUIs. 

% Arguably, certain intents become cumbersome in one sentence, 
% Advising, MS Marco

\begin{table}[t!]
    \small
    \begin{tabularx}{\linewidth}{lL}
         \textit{User:} & \textit{What is the largest 11780 fixed disk under \$40 000?}\\
         \textit{Wizard:} & \textit{The rp07-aa is a 516MB fixed pack disk that costs \$38 000.} \\
         \textit{User:} & \textit{The largest under \$50 000?}\\
         \textit{Wizard:} & \textit{The rp07-aa.} \\
    \end{tabularx}
    \caption{Example taken from \citet{Carbonell1983DiscourseInterfaces}. ``The largest under \$50 000?'' is an elliptical utterance (because the part about 11780 fixed disk is omitted).}
    \label{table:ellipses}
\end{table}

% \begin{table}[t!]
%     \small
%     \begin{tabularx}{\linewidth}{XR}
%          Does the highest paid female manager have any degrees from Harvard? &  \\
%          & Yes, 1. \\
%          How about MIT? & \\
%          & No, none. \\
%          Who is the manager of the largest department? & \\
%          & \begin{tabular}{lll}\textbf{Name} &  \textbf{Dept.} & \textbf{Count}\\ Patterson & 045 & 40\end{tabular}\\
%          The smallest department? & \\
%          & \begin{tabular}{lll} \textbf{Name}  & \textbf{Dept.} & \textbf{Count}\\ Saavedra &  011 &    2\end{tabular}
%     \end{tabularx}
%     \caption{Example taken from \citet{Androutsopoulos1995NaturalIntroduction}. ``How about MIT'' and ``The smallest department`` are examples of elliptical utterances.}
%     \label{table:ellipses}
% \end{table}

\begin{table*}[t]
\centering
\begin{tabular}{lll}
\toprule
Deviation & Project\\
\midrule
\multirow{3}{*}{Synthetic language} & BabyAI~\cite{Chevalier-Boisvert2019BabyAI:Loop} & \\
& CLEVR~\cite{Johnson2017Clevr:Reasoning} & \\
& CFQ~\cite{Keysers2019MeasuringData} & \\
& GQA~\cite{hudson_gqa_2019} & \\
\midrule
\multirow{4}{*}{Artificial task} & GuessWhat~\cite{DeVries2017GuesswhatDialogue} & \\
& CerealBar~\cite{Suhr2019ExecutingInteractions} & \\
& CoDraw~\cite{Kim2019CoDraw:Communication} & \\
& VisionAndLanguage~\cite{Anderson2018Vision-and-LanguageEnvironments} & \\
\midrule
\multirow{4}{*}{Not working with prospective users} & Visual Question Answering~\cite{Antol2015VQA:Answering} \\
& Visual Dialog~\cite{Das2017VisualDialog} & \\
& Spider~\cite{Yu2018Spider:Task} & \\
& SQuAD~\cite{Rajpurkar2016SQuAD:Text} & \\
\midrule
\multirow{5}{*}{Scripts and priming} & MultiWOZ~\cite{Budzianowski2018MultiWozModelling} &\\
& ALFRED~\cite{Shridhar2020Alfred:Tasks} & \\
& CoSQL~\cite{Yu2019CoSQL:Databases} & \\
& Sparc~\cite{Yu2019SParC:Context} & \\
& AirDialogue~\cite{Wei2018AirDialogue:Research} & \\
& Overnight~\cite{Wang2015BuildingOvernight} & \\
\midrule 
\multirow{4}{*}{Single-turn interfaces} & Advising~\cite{Finegan-Dollak2018ImprovingMethodology}\\
& MS Marco~\cite{Bajaj2016MSDataset}\\
& Natural Questions~\cite{Kwiatkowski2019NaturalResearch}\\
& DuReader~\cite{He2018DuReader:Applications}\\
\bottomrule
\end{tabular}
\caption{Five common deviations from the proposed ecologically valid research procedure. For each deviation we list a number of recent LUI benchmarks that suffer from it.}
\end{table*}

\section{Other Ecological Validity Concerns}

Besides the five common deviations, there are two other ecological validity concerns which we did not discuss so far: (i) the evaluation of machine learning models for LUI benchmarks and (ii) the relevance of speech interfaces. 

\paragraph{Evaluation} \label{sec:wrongevaluation} 
Automatic evaluation procedures are key to enable fast iteration of machine learning models. In the context of language user interfaces, practitioners often evaluate their systems with turn-based metrics which, for example, compare the predicted database query to the groundtruth one~\cite{Finegan-Dollak2018ImprovingMethodology} or assess if the simulated robot has behaved in a desired way. This turn-based evaluation procedure assumes that the system followed the ground-truth conversation up to the $(N-1
)^{th}$ turn and then measures the performance for the $N^{th}$ response. The key issue with this evaluation procedure is that it does not account for errors that the system makes along the conversation.\footnote{In the machine translation community, researchers refer to this issue as the ``exposure bias''~\cite{Wiseman2016Sequence-to-SequenceOptimization}.} For example, imagine that the evaluated system makes an error that a human wizard would never make. In the next turn, the user will clarify their intent and thereby diverge to a dialogue that has zero probability under the training distribution (as the wizard would never have made the error). Thus, evaluating under the assumption of ground-truth inputs does not measure how well the system is able to recover from its own mistakes. The only way to measure that is through a human-in-the-loop evaluation that assesses whether the interaction as a whole was successful. 

% Dialog models are closed-loop systems for which the output of the model strongly affects the next user utterance.
% Second, the ultimate goal of this line of research is to improve the user's satisfaction with the natural language interface. Another potential pitfall is thus that the designed evaluation metrics do not align with user satisfaction, leading to the false belief of a better-performing NLI when we are observing an improvement in the metrics. Human-in-the-loop evaluation with the correct metrics are thus critical to ensure one is developing natural language interfaces with positive impact on society. 

% \cmm{But note that texting/messaging is also very widespread and will be readily used in other situations}

\paragraph{Speech interfaces} One aspect that we do not dwell on is the importance of voice-controlled interfaces for the ecological validity of LUI benchmarks. While texting and messaging is very widespread, there are situations in which speech is the preferred interface, such as in settings where people cannot use their hands, e.g., while driving or cooking. Collecting ecologically valid data for such LUI benchmarks will bring additional challenges, including the handling of speech disfluencies, barge-in, and non-verbal cues. We leave these speech-related concerns for future work. 

\section{Directions for Future Research}

Looking forward, there are number of directions that we think deserve more attention from the NLP and ML communities. First, we believe more effort should be put in designing ecologically valid LUI tasks. One approach is to construct LUI tasks for environments that already have many users and which will allow collection of large datasets. Promising proposals are the development of LUI benchmarks for popular video game environments like Minecraft~\cite{Szlam2019WhyMinecraft} or for platforms that bundle user services on the Internet of Things~\cite{Campagna2019Genie:Commands}. A more ambitious direction is to create LUIs that have the potential to attract a big user audience. For example, the academic community could work on LUIs that enable citizens to easily access statistical information published by governments.  

Our second recommendation is that, as a first step, the community could focus on ecologically valid \emph{evaluation}. As collecting large amounts of ecologically valid training data remains expensive, it would be be easier to start with smaller amounts of data for testing purposes. Such an evaluation procedure would directly measure to what extent the trained model generalizes to a practical NLI use case. For training, one could still use data with low ecological validity---e.g., by data augmentation on real language~\cite{Andreas2019Good-enoughAugmentation}---so as to meet the big data requirements of deep learning methods. 

Finally, as many current LUI benchmarks suffer from low ecological validity, we recommend researchers \emph{not} to initiate incremental research projects on them. Benchmark-specific advances are less meaningful when it is unclear if they transfer to real LUI use cases. Instead, we suggest the community to focus on conceptual research ideas that can generalize well beyond the current datasets.

\paragraph{Acknowledgement}
We would like to thank Nils Dahlb\"{a}ck, Nicolas Chapados, Christopher Pal, Siva Reddy,  Torsten Scholak, Raymond Li, Nathan Schucher, and Michael Noukhovitch for helpful discussions on this topic. 

\newpage

\bibliography{harm_mendeley,dima_zotero,chris}
\bibliographystyle{acl_natbib}

\end{document}